\documentclass[preprint,12pt]{elsarticle}

\usepackage{amssymb}
\usepackage[table,xcdraw]{xcolor}
\usepackage{algorithm}
\usepackage{algpseudocode}
\usepackage{caption}

\journal{arXiv}

\begin{document}

\begin{frontmatter}

\title{A multimodal deep learning architecture for smoking detection with a small data approach}

\author[2,3,4]{Róbert Lakatos\corref{cor1}}
\ead{lakatos.robert@inf.unideb.hu}
\cortext[cor1]{Corresponding author}
    
\affiliation[1]{
    organization={Data-Driven Health Division of National Laboratory for Health Security, Health Services Management Training Centre, Semmelweis University},
    addressline={2 Kútvölgyi Road}, 
    city={Budapest},
    postcode={H-1125},
    country={Hungary}}

\affiliation[2]{
    organization={Department of Data Science and Visualization, University of Debrecen, Faculty of Informatics},
    addressline={26 Kassai Road}, 
    city={Debrecen},
    postcode={H-4002},
    country={Hungary}}

\affiliation[3]{
    organization={Doctoral School of Informatics, University of Debrecen},
    addressline={26 Kassai Road}, 
    city={Debrecen},
    postcode={H-4002},
    country={Hungary}}

\affiliation[4]{
    organization={Neumann Technology Platform, Neumann Nonprofit Ltd.},
    addressline={8 Naphegy Square}, 
    city={Budapest},
    postcode={H-1016},
    country={Hungary}}
    
\author[1]{Péter Pollner}
\author[2]{András Hajdu}
\author[1,4]{Tamás Joó}

\begin{abstract}
\textbf{Introduction:} Covert tobacco advertisements often raise regulatory measures. This paper presents that artificial intelligence, particularly deep learning, has great potential for detecting hidden advertising and allows unbiased, reproducible, and fair quantification of tobacco-related media content.
\\
\textbf{Methods:} We propose an integrated text and image processing model based on deep learning, generative methods, and human reinforcement, which can detect smoking cases in both textual and visual formats, even with little available training data.
\\
\textbf{Results:} Our model can achieve 74\% accuracy for images and 98\% for text. Furthermore, our system integrates the possibility of expert intervention in the form of human reinforcement.
\\
\textbf{Conclusions:} Using the pre-trained multimodal, image, and text processing models available through deep learning makes it possible to detect smoking in different media even with few training data.
\end{abstract}

\begin{keyword}
AI supported preventive healthcare \sep pre-training with generative AI  \sep  multimodal deep learning  \sep  automated assessment of covert advertisement  \sep  few shot learning \sep smoking detections
\end{keyword}

\end{frontmatter}

\section{Introduction}
\label{intro}
The WHO currently estimates that smoking causes around 8 million deaths a day. It is the leading cause of death from a wide range of diseases, for example, heart attacks, obstructive pulmonary disease, respiratory diseases, and cancers. 15\% of people aged 15 years and over smoke in the OECD countries and 17\% in the European Union \cite{OECD_Daily_smokers}. Moreover, of the 8 million daily deaths, 15\% result from passive smoking \cite{WHO_smoking_death}. The studies \cite{chapman1997smoking, pechmann1996smoking} below highlight the influence of smoking portrayal in movies and the effectiveness of health communication models. However, quantifying media influence is complex. For internet media like social sites, precise ad statistics are unavailable. Furthermore, calculating incited and unmarked ads poses a significant difficulty as well. Therefore, accurate knowledge of the smoking-related content appearing in individual services can be an effective tool in reducing the popularity of smoking. Methods for identifying content include continuous monitoring of advertising intensity \cite{kong2022understanding}, structured data generated by questionnaires \cite{fielding2004declines}, and AI-based solutions that can effectively support these goals. The authors of the article "Machine learning applications in tobacco research" \cite{fu2023machine} point out in their review that artificial intelligence is a powerful tool that can advance tobacco control research and policy-making. Therefore, researchers are encouraged to explore further possibilities.

Nonetheless, these methods are highly data-intensive. In the case of image processing, an excellent example of this is the popular ResNet \cite{he2016deep} image processing network, which was trained on the ImageNet dataset \cite{deng2009imagenet} containing 14,197,122 images. Regarding text processing, we can mention the popular and pioneering BERT network \cite{devlin2018bert} trained by the Toronto BookCorpus \cite{zhu2015aligning} was trained by the 4.5 GB of Toronto BookCorpus. Generative text processing models such as GPT \cite{radford2018improving} are even larger and were trained with significantly more data than BERT. For instance, the training set of GPT 3.0 was the CommonCrawl \cite{commonCrawl} dataset, which has a size of 570 GB.

The effective tools for identifying the content of natural language texts are topic modeling \cite{blei2003latent} and the embedding of words \cite{pennington2014glove, mikolov2013distributed, bojanowski2017enriching}, tokens, sentences \cite{reimers2019sentence}, or characters \cite{clark2022canine} clustering \cite{arthur2006k}. For a more precise identification of the content elements of the texts, we can use the named-entity recognition \cite{ali2022named} techniques. In image processing, we can highlight classification and object detection to detect smoking. The most popular image processing models are VGG \cite{simonyan2014very}, ResNet \cite{he2016deep}, Xception \cite{chollet2017xception}, EfficientNet \cite{tan2019efficientnet}, Inception \cite{szegedy2016rethinking}, and YOLO \cite{redmon2016you}. Moreover, there are architectures like CAMFFNet \cite{lin2022camffnet}, which are specifically recommended for smoking detection. The development of multimodal models also is gaining increasing focus \cite{liu2019cyclematch, liu2022image}, which can use texts and images the solve the tasks at the same time. For movies, scene recognition is particularly challenging compared to images \cite{rao2020local}.  Scene recognition is also linked to sensitive events such as fire, smoke, or other disaster detection systems \cite{gagliardi2021real}, but there are attempts to investigate point-of-sale and tobacco marketing practices \cite{bianco2021automated} as well.

We concluded that there is currently no publicly available specific smoking-related dataset that would be sufficient to train a complex model from scratch. Hence, we propose a multimodal architecture that uses pre-trained image and language models to detect smoking-related content in text and images. By combining image processing networks with multimodal architectures and language models, we leverage textual and image data simultaneously. This offers a data-efficient and robust solution that can be further improved with expert input. This paper demonstrates the remarkable potential of artificial intelligence, especially deep learning, for the detection of covert advertising, alongside its capacity to provide unbiased, replicable, and equitable quantification of tobacco-related media content.

\clearpage

\section{Methods}

\subsection{Model Architecture} 
\label{model_architecture}

\begin{figure}[!ht]
  \begin{center}        
    \includegraphics[width=340pt]{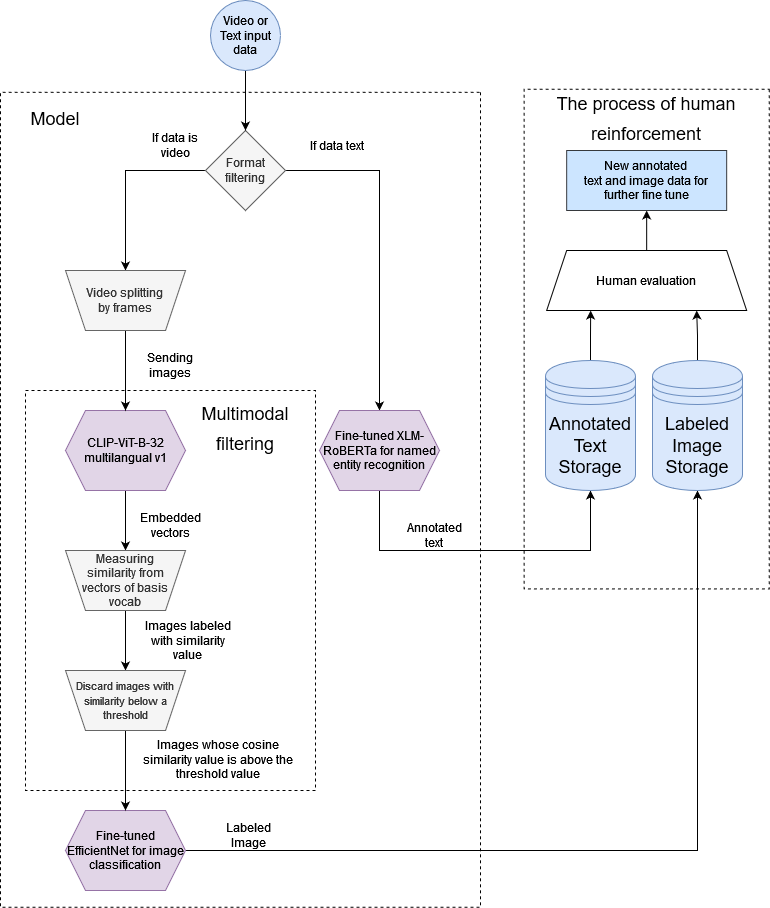}
  \end{center}
  \caption{Schematic flow diagram of the architecture.}
  \label{fig:flow_diagram}
\end{figure}

As illustrated in Figure \ref{fig:flow_diagram} by a schematic flow diagram, our solution relies on pre-trained language and image processing models and can handle both textual and image data.  

The first step of our pipeline is to define the incoming data format because need to direct the data to the appropriate model for its format. The video recordings are analyzed with multimodal and image processing models, while the texts are analyzed with a large language model. In the case of video recordings, we applied the CLIP-ViT-B-32 multilingual \cite{reimers2020making, radford2021learning} model. The model has been developed for over 50 languages with a special training technique \cite{reimers2020making}. The model supports Hungarian, which was our target language. We use the CLIP-ViT-B-32 model as a filter. After filtering, to achieve more accurate results, we recommend using the pre-trained EfficientNet B5 model, which we fine-tuned with smoking images for the classification task. 

To process texts, we use name entity recognition to identify smoking-related terms. For this purpose, we have integrated into our architecture an XLM-RoBERTa model \cite{DBLP:journals/corr/abs-1911-02116} that is pre-trained, multilingual, and also supports the Hungarian language, which is important to us.

\subsection{Format check} 
\label{format_check}

The first step in processing is deciding whether the model has to process video recordings or text data. Since there are many formats for videos and texts, we chose the simple solution of only supporting mp4 and txt file formats. The mp4 is a popular video format, and practically all other video recording formats can be converted to mp4. We consider txt files utf8-encoded raw text files that are ideally free of various metadata. It is important to emphasize that here we ignore the text cleaning processes required to prepare raw text files. The reason is that we did not deal with faulty or txt files requiring further cleaning during the trial.

\subsection{Processing of videos and images} 
\label{process_video_img}

The next step in the processing of processing video footage is to break it down into frames by sampling every second. The ViT image encoder of the CLIP-ViT-B-32 model was trained by its creators for various image sizes. For this, they used the ImageNet \cite{deng2009imagenet} dataset in which the images have an average size of 469$\times$387 pixels.

The developers of CLIP-ViT-B-32 do not recommend an exact resolution for the image encoder. The model specification only specifies a minimum resolution of 224$\times$224.  In the case of EfficientNetB5, the developers have optimized an image size of 224$\times$224. For these reasons, we have taken this image size as a reference and transformed the images sampled from the video recordings to this image size.

\subsection{Multimodal filtering} 
\label{multimodal_filtering}

The images sampled from the video recordings were filtered using the CLIP-ViT-B-32 multilingual v1 model. The pre-trained CLIP-ViT-B-32 multilingual v1 model consists of two main components from a ViT \cite{dosovitskiy2020image} image processing model and a DistilBERT-based \cite{Sanh2019DistilBERTAD} multilingual language model.  We convert into a 512-long embedded vector \cite{mikolov2013distributed} the images and texts with CLIP-ViT-B-32. The embedded vectors for texts and images can be compared based on their content meaning if we measure cosine similarities between the vectors. The cosine similarity is a value falling in the interval [-1,1], and the similarity of two vectors will be larger the closer their cosine similarity is to 1.

Since we aimed to find smoking-related images, we defined a smoking-related term. We converted it to a vector and measured it against the embedded vectors generated from the video images. The term we chose was the word "smoking". We can use more complex expressions, which could complicate the measurement results interpretation.

The cosine similarity of the vectors produced by embedding the images always results in a scalar value compared to the vector created from our expression related to "smoking".  However, the decision limit between the distances measured between the vectors produced by the CLIP-ViT-B-32 model is not always clear. Namely, even in the case of images with meanings other than "smoking", we get a value that is not too distant.

\begin{figure}[!ht]
  \begin{center}          
    \includegraphics[width=280pt]{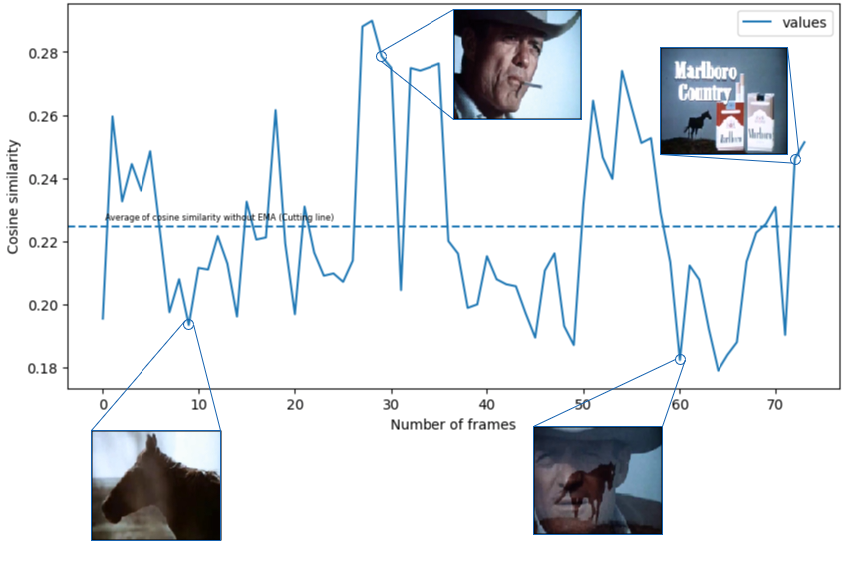}
  \end{center}
  \caption{The cosine similarity of the images obtained from the video recording in chronological order.}
  \label{fig:signal_0}
\end{figure}

We had to understand the distribution of the smoking images to eliminate this kind of blurring of the decision boundary. To this end, we examined the characteristics of the distribution of the images. It is clear from Figure \ref{fig:signal_0} that because the images with a semantic meaning closer to smoking appear randomly in a video recording, it is difficult to grasp the series of images that can be useful for us.  Figure \ref{fig:signal_0} is actually a function whose vertical axis has the cosine similarity values belonging to the individual images. At the same time, the horizontal axis shows the position of the images in the video. To solve this problem, we introduced the following procedure. If we put the cosine similarity values in ascending order, we get a function that describes the ordered evolution of the cosine similarity values. 

\break

\begin{figure}[!ht]
  \begin{center}
    \includegraphics[width=280pt]{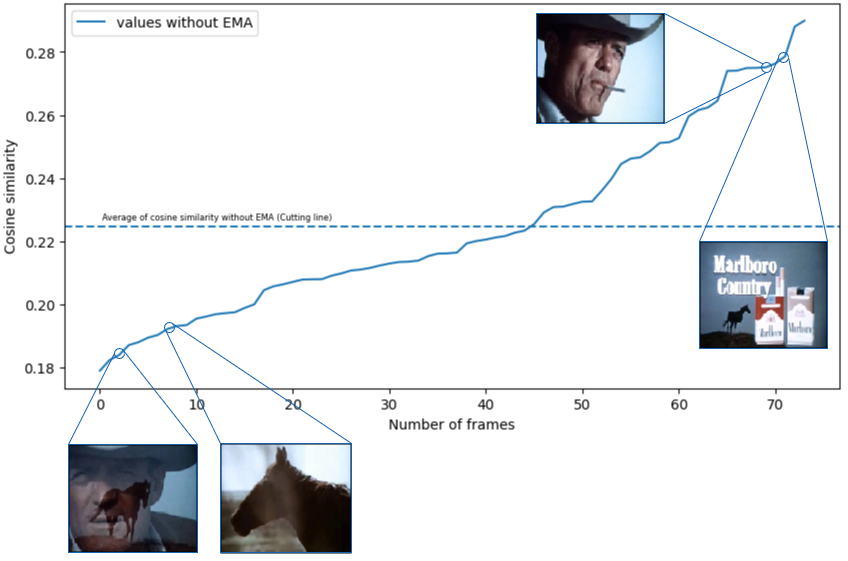}
  \end{center}
  \caption{The images are in an orderly manner based on the cosine similarity values.}
  \label{fig:signal_1}
\end{figure}

The ordered function generated from Figure \ref{fig:signal_0} can be seen in Figure \ref{fig:signal_1}. As shown in Figures \ref{fig:signal_0} and \ref{fig:signal_1}, we found that if we take the similarity value of the images sampled from the given sample to the word "smoking", their average results in a cutting line, and we can use it as a filter.

\break

Furthermore, considering the specifics of the video recordings, we consider that the average can be corrected with a constant value. In this mean, the constant value can thus also be defined as the hyperparameter of the model. We chose the 0 default value for the correction constant because of more apparent measurements. Because the choice of the best constant value may differ depending on the recording type and may distort the exact measurement results. 

\subsection{Fine-tuned image classification} 
\label{efficentnet_classif}

After filtering the image set with a multimodal model, we applied an image processing model to classify the remaining images further to improve accuracy. Among the publicly available datasets on smoking, we have used the "smoker and non-smoker" \cite{khan2020dataset} for augmented \cite{shorten2019survey} fine-tuning. We selected the following models for the task. EfficientNet, Inception, ResNet, VGG, and Xception. The EfficientNet B5 version was the best, with an accuracy of 93.75\%. Table S1 of the supplemental contains our detailed measurement results concerning all models.

\subsection{Processing of text} 
\label{text}

In the case of detecting smoking terms in texts, we approached the problem as an NER task and focused on the Hungarian language. Since we could not find a dataset containing annotated smoking phrases available in Hungarian. Therefore, to generate the annotated data, we used the generational capabilities of ChatGPT, the smoking-related words of the Hungarian synonyms and antonyms dictionary \cite{violet-opposites}, and prompt engineering. Accordingly, we selected words related to smoking from the synonyms and antonyms dictionary and asked ChatGPT to suggest further smoking-related terms besides words from the Hungarian dictionary. Finally, we combined the synonyms and the expressions generated by ChatGPT into a single dictionary.

We created blocks of a maximum of 5 elements from the words in our dictionary. Each block contained a random combination of a maximum of 5 words. The blocks are disjoint, so they do not contain the same words. This mixing step was done 10 times. This means that, in one iteration, we could form 8 blocks of 5-element disjunct random blocks from our 43-word dictionary. By doing all these 10 times, we produced 80 blocks. However, due to the 10 repetitions, the 80 blocks were no longer disjoint. In other words, if we string all the blocks together, we get a dictionary in which every synonym for smoking appears a maximum of 10 times. 

We made a prompt template to which, by attaching each block, we instructed ChatGPT to generate texts containing the specified expressions. Since ChatGPT uses the Hungarian language well, the generated texts contained our selected words by the rules of the Hungarian language, with the correct conjugation. An example of our prompts is illustrated in Table \ref{table:prompt}.

\begin{table}[!ht]
\footnotesize
\begin{center}
\caption{A 3 elements example prompt for ChatGPT.}
\begin{tabular}{|c|}
\hline
\begin{tabular}[c]{@{}c@{}}Generate a short text about smoking. \\ The text strictly contains the following words in the different sentences: \\ {\color[HTML]{6665CD} smoking}, {\color[HTML]{00D2CB} tobacco}, {\color[HTML]{F56B00} cigar} \end{tabular} \\ \hline
\end{tabular}
\label{table:prompt}
\end{center}
\end{table}

We did not specify how long texts should be generated by ChatGPT or that every word of a 5-element block should be included in the generated text. When we experimented with ChatGPT generating fixed-length texts, it failed. Therefore, we have removed the requirement for this. Using this method, we created a smoking-related corpus consisting of 80 paragraphs, 49000 characters, and 7160 words. An English example of a generated text is presented in Table 2.

\begin{table}[!ht]
\footnotesize
\begin{center}
\caption{An example paragraph generated by from the prompt of Table \ref{table:prompt}.}
\begin{tabular}{|c|}
\hline
\begin{tabular}[c]{@{}c@{}}
{\color[HTML]{6665CD}Smoking} is a widespread and addictive habit that involves inhaling \\
and exhaling the {\color[HTML]{6665CD}smoke} produced by burning {\color[HTML]{00D2CB}tobacco}. Whether it's \\
a hand-rolled {\color[HTML]{F56B00}cigar} or a manufactured {\color[HTML]{F56B00}cigarette}, the act of \\
smoking revolves around the consumption of {\color[HTML]{00D2CB}tobacco}. Despite the well-known \\
health risks, many individuals continue to engage in smoking due \\
to its addictive nature. The allure of a {\color[HTML]{F56B00}cigar} or \\
a {\color[HTML]{F56B00}cigarette} can be strong, making it challenging for people \\
to quit smoking even when they are aware of its \\
detrimental effects. Education and support are crucial in helping individuals \\
break free from the cycle of {\color[HTML]{6665CD}smoking} and its associated \\
harms. \\
\end{tabular} \\ \hline
\end{tabular}
\label{table:paragraph}
\end{center}
\end{table}

To find the best model according to the possibilities of our computing environment and the support of the Hungarian language, we tested the following models: XLM RoBERTa base and large, DistilBERT base cased, huBERT base \cite{Nemeskey:2021a}, BERT base multilingual \cite{bert-multilingual}, Sentence-BERT \cite{reimers-2019-sentence-bert}. The best model was the XLM RoBERTa large one, which achieved 98\% accuracy and 96\% F1-score on the validation dataset and an F1-score of 91\% with an accuracy of 98\% on the test dataset.

\subsection{Human reinforcement} 
\label{human_reinf}

In the architecture we have outlined, the last step in dealing with the lack of data is to ensure the system’s continuous development capability. For this, we have integrated human confirmation into our pipeline. The essence is that our system's hyperparameters should be adjustable and optimizable during operation and that the data generated during detection can be fed back for further fine-tuning. The cutting line used in multimodal filtering is a hyperparameter of our model. As a result, a more accurate result can be achieved by using human confirmation during the operation. The tagged images and annotated texts from the processed video recordings and texts are transferred to permanent storage in the last step of the process. This dynamically growing dataset can be further validated with additional human support, and possible errors can be filtered. So, False positives and False negatives can be fed back into the training datasets.

\section{Results} 
\label{results}

We collected video materials to test the image processing part of our architecture. The source of the video materials was the video-sharing site YouTube. Taking into account the legal rules regarding the usability of YouTube videos, we have collected 5 pieces short advertising films from the Malboro and Philip Moris companies. We ensured not to download videos longer than 2 minutes because longer videos, such as movies, would have required a special approach and additional pre-processing. Furthermore, we downloaded the videos at 240p resolution and divided them into frames by sampling every second. Each frame was transformed to a resolution of 224×224 pixels. We manually annotated all videos. The downloaded videos averaged 64 seconds and contained an average of 13 seconds of smoking. 

With the multimodal filtering technique, we discarded the images that did not contain smoking. Multimodal filtering found 25 seconds of smoking on average in the recording. The accuracy of the identified images was 62\%. The multimodal filtering could filter out more than half of the 64-second, on average, videos. We also measured the performance of the fine-tuned EfficientNet B5 model by itself. The model detected an average of 28 seconds of smoking with 60\% accuracy. We found that the predictions of the two constructions were sufficiently diverse to connect them using the boosting ensemble \cite{dietterich2000ensemble} solution. By connecting the two models, the average duration of perceived smoking became 12 seconds with 4 seconds on average error and 74\% accuracy.  The ensemble solution was the best approach since the original videos contained an average of 13 seconds of smoking. We deleted the videos after the measurements and did not use them anywhere for any other purpose.

We created training and validation datasets from Hungarian synonyms for smoking using ChatGPT. We trained our chosen large language models until their accuracy on the validation dataset did not increase for at least 10 epochs. The XLM-RoBERTa model achieved the best performance on the validation dataset with an F1-score of 96\% and 98\% accuracy. For the final measurement, we created test data from an online text related to smoking by manual annotation \cite{semmelweis}. The text of the entire test data is included in the Table S20 supplemental. The fine-tuned XLM-RoBERTa model achieved 98\% accuracy and 0.91 F1 score on the test dataset.

\section{Conclusions} 
\label{conclusion}

Multimodal and image classification models are powerful for classification tasks. In return, however, they are complex and require substantial training data, which can reduce their explainability and usability. In turn, our solution showed that pre-trained multimodal and image classification models exist that allow smoking detection even with limited data and in the matter of low-resource languages if we use the potential of human reinforcement, generative, and ensemble methods. In addition, we see further development opportunities if our approach is supplemented with an object detector, which can determine the time of occurrence of objects and their position. Moreover, with the expected optimization of the automatic generation of images in the future and the growth of the available computing power, our method used for texts can work in the case of images.

\clearpage

\section*{Funding}

The project no. KDP-2021 has been implemented with the support provided by the Ministry of Culture and Innovation of Hungary from the National Research, Development, and Innovation Fund, financed under the C1774095 funding scheme. Also, this work was partly funded by the project GINOP-2.3.2-15-2016-00005 supported by the European Union, co-financed by the European Social Fund, and by the project TKP2021-NKTA-34, implemented with the support provided by the National Research, Development, and Innovation Fund of Hungary under the TKP2021-NKTA funding scheme.  In addition, the study received further funding from the National Research, Development and Innovation Office of Hungary grant (RRF-2.3.1-21-2022-00006, Data-Driven Health Division of National Laboratory for Health Security).

\bibliographystyle{elsarticle-num} 
\bibliography{cas-refs}

\end{document}